%% file: root.tex
\documentclass[letterpaper, 10 pt, conference]{ieeeconf}  

\IEEEoverridecommandlockouts                              

\input{commands}
\usepackage{graphics} 
\usepackage{epsfig} 
\usepackage{amsmath}
\usepackage{amssymb}  
\usepackage[ruled, vlined, titlenotnumbered, linesnumbered]{algorithm2e}
\usepackage{multirow}
\usepackage{tabularx}
\usepackage{booktabs}
\usepackage{caption}
\usepackage{bbding}

\usepackage{graphicx}
\usepackage{longtable}
\usepackage{array}

\usepackage{multicol}
\usepackage{float}
\usepackage{tablefootnote}
\usepackage{pifont}
\usepackage{wrapfig}
\newcommand{\cmark}{\ding{52}}%
\newcommand{\xmark}{\ding{55}}%

\makeatletter
\let\NAT@parse\undefined
\makeatother
\usepackage[colorlinks, citecolor=green]{hyperref}

\usepackage{subcaption}
\usepackage[table]{xcolor}
\usepackage{soul}
\usepackage[normalem]{ulem}

\hypersetup{
    colorlinks=true,
    linkcolor=blue,
    filecolor=magenta,      
    urlcolor=blue,
    pdftitle={SSI-Policy},
    pdfpagemode=FullScreen,
    }

\title{\LARGE \bf
SSI-Policy: Learning Structured Scene Interfaces for Vision-Language Robotic Manipulation
}

\author{
Kaijun Wang$^{\dagger 1}$, 
Zikai Ouyang$^{\dagger 1,2}$, 
Xuping Wu$^{1}$, 
Jinyi Hong$^{1}$, 
Wei Pan$^{1}$, 
Haibo Lu$^{2}$, 
Jia Pan$^{3}$, \\
Wei Zhang$^{\ddagger 1,4}$, and 
Linfang Zheng$^{\ddagger 3}$
\thanks{*This work was supported by the Natural Science Foundation of China (62461160309), the NSFC-RGC Joint Research Scheme (N\_HKU705/24), Hong Kong RGC (GRF 17201025, GRF17200924), the National Science and Technology Major Project (2024ZD01NL00100), and the Major Key Project of PCL (PCL2025A03).}
\thanks{$^\dagger$denotes equal contribution, $^\ddagger$denotes corresponding authors.}
\thanks{$^{1}$Kaijun Wang, Zikai Ouyang, Xuping Wu, Jinyi Hong, Wei Pan, and Wei Zhang are with the Southern University of Science and Technology, Shenzhen 518055, China.
{\tt\small \{12232628, ouyang2022, wuxp, 12213005, 12211810\}@mail.sustech.edu.cn, zhangw3@sustech.edu.cn}}%
\thanks{$^{2}$Zikai Ouyang and Haibo Lu are with Peng Cheng Laboratory, Shenzhen 518000, China.
{\tt\small luhb@pcl.ac.cn}}%
\thanks{$^{3}$Jia Pan and Linfang Zheng are with The University of Hong Kong, Hong Kong SAR, China.
{\tt\small jpan@cs.hku.hk, lfzheng@hku.hk}}%
\thanks{$^{4}$Wei Zhang is also with LimX Dynamics, China.}%
}

\begin{document}

\maketitle
\thispagestyle{empty}
\pagestyle{empty}

\input{tex/abstract}
\input{tex/intro}
\input{tex/related}
\input{tex/method}
\input{tex/experiments}
\input{tex/conclusion}








\bibliographystyle{IEEEtran}
\bibliography{references}

\end{document}

%% file: commands.tex
\usepackage{bm}
\newcommand{\inputRGBs}{\mathbf{I}}
\newcommand{\RGB}{I}
\newcommand{\proprioSeq}{\mathbf{P}}
\newcommand{\proprio}{p}
\newcommand{\actSeq}{\mathbf{A}}
\newcommand{\action}{a}
\newcommand{\policy}{\pi}
\newcommand{\textInstruct}{\ell}
\newcommand{\spatialSignals}{\mathbf{S}}
\newcommand{\perceptNet}{f_{\text{percept}}}
\newcommand{\policyNet}{f_{\text{action}}}
\newcommand{\trajectories}{\mathcal{T}}
\newcommand{\depthImage}{D}
\newcommand{\bboxMap}{B}
\newcommand{\singleTraj}{\tau}
\newcommand{\pixel}{\mathbf{x}}
\newcommand{\trajStepLen}{H'}

\newcommand{\fusedSpatialFeat}{s}
\newcommand{\obsStep}{t_o}

\newcommand{\denoisingStep}{k}

%% file: tex/abstract.tex
\begin{abstract}
Real-world robotic manipulation demands spatial grounding, task-aware reasoning, and precise control. Learning such capabilities becomes particularly challenging in the low-data regime. Prior methods often trade off scalable task-level reasoning and explicit physical structure: video-based approaches can drift geometrically over long horizons, 3D approaches often require depth sensing, and many flow/trajectory interfaces emphasize motion without an explicit RGB-only geometric representation. We introduce SSI-Policy, a modular framework built around a Structured Scene Interface (SSI)---a unified, RGB-only intermediate representation that jointly encodes monocular depth features, language-grounded object layouts, and instruction-conditioned 2D motion trajectories. Critically, SSI is robot-agnostic and trainable from action-free video, decoupling perception from control so that the downstream policy can learn from few demonstrations. On the LIBERO benchmark with only 10 demonstrations per task, SSI-Policy improves over the strongest prior method by nearly 15\% and remains competitive with 50-demo methods that leverage large-scale external pretraining. Ablations show that geometric and motion cues provide complementary benefits within the shared interface. We further validate on 13 real-world tasks spanning spatial reasoning, cross-embodiment transfer, and contact-rich manipulation.
\end{abstract}

%% file: tex/intro.tex
\section{Introduction}
\label{sec:intro}

Real-world robotic manipulation requires more than semantic recognition—it demands spatial grounding, task-aware reasoning, and precise control. A robot must understand not only what objects are present, but where they are, how they relate geometrically, and how they are expected to move under a given instruction. These requirements become even more challenging in language-conditioned settings and under limited demonstration data, where both spatial ambiguity and data scarcity can significantly degrade policy performance.

Existing approaches often trade off between scalable task-level reasoning and explicit physical structure. Video-based generative methods plan by forecasting future observations~\cite{UniPi_Du_NIPS_2023, baker2022video, Dreamitate_Liang_CoRL_2024}; while they benefit from large-scale RGB pretraining, operating in pixel space makes long-horizon rollouts computationally expensive and prone to geometric drift. Geometry-centric methods improve spatial reasoning via 3D representations~\cite{DepthHelps_2024_IROS, SpatialVLA_Qu_2025, PerAct_Shridhar_CORL_2022, VisualRoboticManipulationDepthAware_Wang_2024}, but often rely on RGB-D sensing and struggle in situations where multi-task language-conditioned settings are required. Recent flow/trajectory-based interfaces provide compact motion abstractions, yet each carries notable limitations: ATM~\cite{ATM_Wen_RSS_2024} treats geometry implicitly; Im2Flow2Act~\cite{Im2Flow2Act_Xu_CoRL_2024} and GeneralFlow~\cite{GeneralFlow_yuan2024} require 3D anchor initialization from depth or point-cloud observations; and FLIP~\cite{FLIP_gao2025} relies on flow-conditioned video generation with beam-search rollouts at test time.

\input{images/teaser/item}

Despite these advances, to our knowledge, no existing method provides a unified, RGB-only intermediate interface that jointly encodes geometric structure and task-conditioned motion while remaining embodiment-agnostic and trainable without robot action labels---properties critical for data-efficient multi-task manipulation.

In this work, we introduce \textbf{SSI-Policy}, a modular framework built around a \textbf{Structured Scene Interface (SSI)}—a unified intermediate representation that explicitly encodes geometric structure and task-conditioned motion using only RGB inputs. The key idea is not merely to combine useful features, but to define a principled interface layer between perception and control: one that is robot-agnostic, trainable from action-free RGB videos, and consumable by a lightweight policy with minimal demonstrations. Within this interface, SSI jointly models three complementary signals: (i) monocular depth features capturing relative scene geometry, (ii) language-grounded object layout maps highlighting task-relevant regions, and (iii) instruction-conditioned 2D motion trajectories modeling anticipated object and manipulator dynamics. This enables the policy to reason explicitly about \emph{what} to act on, \emph{where} to act, and \emph{how} to move—without depth sensors or test-time video generation.

Crucially, SSI abstracts visual understanding into a structured, task-aligned spatial interface that is independent of embodiment-specific actions. This design allows perception modules to be pretrained from action-free RGB videos (without robot action labels), while the downstream policy is learned from only a small number of task demonstrations. By decoupling perception from control in this way, SSI-Policy improves data efficiency and facilitates cross-embodiment transfer.

We empirically validate this design on the LIBERO benchmark and real-world tasks. With only 10 demonstrations per task, SSI-Policy achieves leading performance in the few-shot setting among methods. The structured interface further supports cross-embodiment transfer, including human-to-robot scenarios. When trained with more demonstrations, SSI-Policy scales competitively with approaches that leverage large-scale external pretraining, despite being trained solely on LIBERO.

Our contributions are threefold:
(i) We propose \textbf{SSI}, a robot-agnostic intermediate interface that jointly encodes monocular geometry, language-grounded layouts, and instruction-conditioned motion from RGB alone, and show through ablations that these signals provide complementary benefits that neither achieves in isolation;
(ii) We demonstrate that this interface design enables data-efficient multi-task learning and cross-embodiment transfer by decoupling perception (trainable from action-free video) from control; and
(iii) We validate SSI-Policy on LIBERO~\cite{LIBERO_Liu_2023} and 13 real-world tasks, achieving strong few-shot performance without depth sensors or external pretraining data.

%% file: images/teaser/item.tex
\begin{figure}[!t]
    \centering
    \includegraphics[width=1\linewidth, trim = 100 110 100 110, clip]{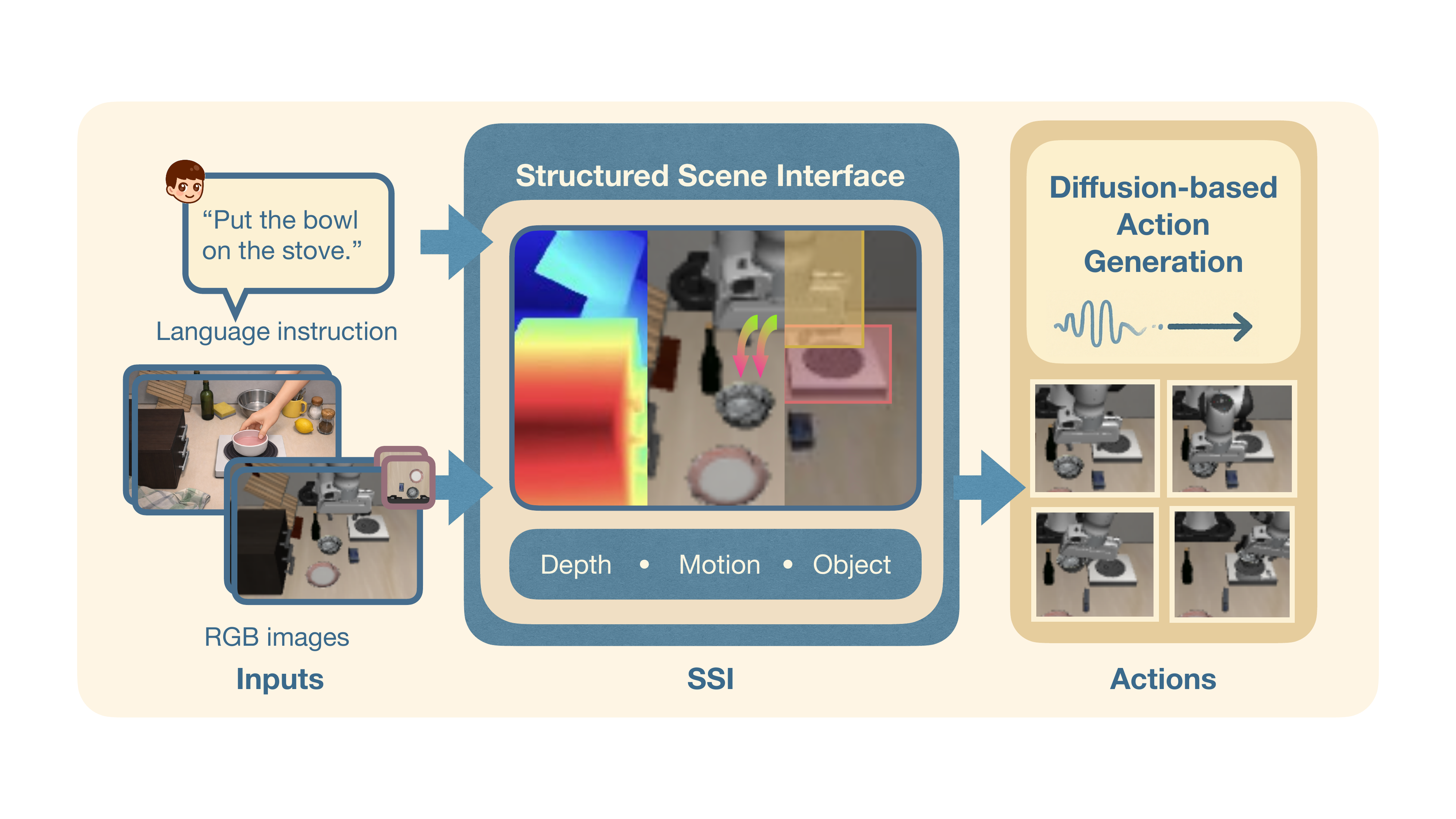}
    \vspace{-20px}
    \caption{\footnotesize \textbf{Overview of SSI-Policy.} From RGB observations and a language instruction, SSI constructs a structured intermediate interface that encodes monocular geometry cues, language-grounded object layouts, and instruction-conditioned motion trajectories. SSI is robot-agnostic (cross-embodiment) and can be learned from action-free videos alone (e.g., human-hand videos or robot videos). A diffusion policy is then trained from few robot demonstrations to generate actions conditioned on SSI, enabling data-efficient manipulation without depth sensors.}
    \label{fig:teaser}
    \vspace{-16px}
\end{figure}

%% file: tex/related.tex
\section{Related Work}
\label{sec:related}

\paragraph{Video-Based Methods for Manipulation}
Video prediction has been explored as a way to model manipulation dynamics. Implicit approaches~\cite{ILPO_edwards2019_ICML, LAPO_schmidt2024_ICLR, Genie_bruce2024_ICML, LAPA_ye2025, UniVLA_RSS_25} learn latent action representations via video reconstruction, but require large datasets and produce latent spaces whose connection to executable control is often indirect. Explicit generative methods~\cite{UniPi_Du_NIPS_2023, Dreamitate_Liang_CoRL_2024, baker2022video, GVFTAPE_CoRL_2025, V2A_2025, AVDC_2023} instead generate future frames conditioned on task descriptions to guide control, enabling high-level reasoning but incurring substantial computational cost and accumulating temporal drift or geometric inconsistencies over long horizons. FLIP~\cite{FLIP_gao2025} further performs beam-search-based planning using flow-conditioned video generation, relying on repeated generative rollouts during planning. These approaches highlight the promise of video-based imagination, yet maintaining geometric consistency and efficiency remains challenging.

\paragraph{Flow-Based Representations for Policy Learning}
To reduce reliance on full-frame video generation, recent works adopt flow or pixel trajectories as compact motion abstractions. ATM~\cite{ATM_Wen_RSS_2024} and Im2Flow2Act~\cite{Im2Flow2Act_Xu_CoRL_2024} use dense 2D motion cues to guide policy learning without explicit model-based planning. Im2Flow2Act further assumes explicit initialization of object 3D surface keypoints (absolute anchors) from depth or point-cloud observations and models motion relative to these anchors. SKIL~\cite{SKIL_wang2025} extends starting-point selection using semantic keypoints extracted from RGB-D inputs, while GeneralFlow~\cite{GeneralFlow_yuan2024} learns a 3D flow interface from RGB-D human videos for cross-embodiment transfer. Although these methods improve efficiency by operating in motion space, they either rely on explicit 3D sensing/initialization or treat geometry implicitly. In contrast, our approach explicitly encodes geometric structure together with task-conditioned motion within a unified intermediate interface under RGB-only sensing.

\paragraph{Spatial Representations for Manipulation}
Effective manipulation requires accurate spatial understanding of objects, contacts, and constraints. Explicit 3D approaches encode geometry using voxel~\cite{PerAct_Shridhar_CORL_2022, C2F-ARM_James2022} or point-cloud representations~\cite{Lift3D_Li_CVPR_2023, FrameMining_Liu_2022, PolarNet_Chen_2023, DP3_Ze_RSS_2024, RISE_Wang_2024, Act3D_Gervet_2023}, enabling fine-grained spatial reasoning but typically requiring accurate depth sensing. Other methods recover geometry from RGB via depth prediction or large-scale supervision~\cite{SpatialVLA_Qu_2025, zhen20243dvla3dvisionlanguageactiongenerative, VisualRoboticManipulationDepthAware_Wang_2024}, trading sensing requirements for additional data and training complexity. In contrast, large-scale imitation learning approaches~\cite{RT1_Brohan_2022, RT2_Brohan_2023, Octo_2024, OpenVLA_Kim_2024, BCZ_Jang_ACRL_2021, reed2022generalist, open_x_embodiment_rt_x_2023, liu2023instructionfollowingagentsmultimodaltransformer, pi0_Black_2024, driess2023palmeembodiedmultimodallanguage, OTTER_2025} learn spatial priors implicitly from RGB, but often require massive datasets and offer limited interpretability. Finally, 2D-centric methods based on keypoints or part-level attention~\cite{CoPa_Huang_IROS_2024, ReKep_Huang_2024, RoboABC_Ju_2024, AffKp_RAL_2021} provide lightweight spatial cues, but may lack explicit geometric grounding for contact-rich manipulation.

\textbf{Our method addresses this gap} by defining a unified intermediate representation that jointly encodes geometric structure and task-conditioned motion under RGB-only sensing. Unlike flow-based methods that treat geometry implicitly or require 3D initialization~\cite{ATM_Wen_RSS_2024, Im2Flow2Act_Xu_CoRL_2024, GeneralFlow_yuan2024}, and unlike video-based planners that incur heavy generation cost~\cite{FLIP_gao2025}, our interface preserves geometric structure without depth sensors while remaining computationally efficient and robot-agnostic---enabling both data-efficient learning and cross-embodiment transfer.

%% file: tex/method.tex
\section{Methodology}
\label{sec:method}
\subsection{Problem Formulation}
We address the problem of vision- and language-conditioned robotic manipulation, where the agent predicts a sequence of future actions given recent sensory observations and a natural language instruction. At each timestep $t$, the agent receives:
(i) a sequence of multi-view RGB frames $\inputRGBs_t = \{ \RGB_{t-\obsStep+1}^{1:V}, \dots, \RGB_t^{1:V} \}$ from $V$ cameras, 
(ii) proprioceptive states $\proprioSeq_t = \{ \proprio_{t-\obsStep+1}, \dots, \proprio_t \}$, and 
(iii) a task instruction $\textInstruct$, where $\obsStep$ denotes the temporal window size of the model input. 
The goal is to generate a horizon-$H$ action sequence $\actSeq_t = \{ \action_t, \dots, \action_{t+H-1} \}$, with the policy defined as:
\vspace{-2mm}
\begin{equation}
    \policy: (\inputRGBs_t, \proprioSeq_t, \textInstruct) \rightarrow \actSeq_t.
\vspace{-3mm}
\end{equation}

\input{images/overall_framework/item}
\subsection{Framework Overview}

We propose \textbf{SSI-Policy}, a modular framework that separates perception from control via a mid-level representation termed the \textbf{Structured Scene Interface (SSI)}. As shown in Figure~\ref{fig:framework}, the policy is decomposed into a perception module and an action planner:
\vspace{-1mm}
\begin{equation}
\vspace{-1mm}
    \spatialSignals_t = \perceptNet(\inputRGBs_t, \textInstruct), \quad \actSeq_t = \policyNet(\spatialSignals_t, \inputRGBs_t, \proprioSeq_t),
\end{equation}
where $\perceptNet$ (the \textbf{Perception Composer}) extracts the SSI $\spatialSignals_t$ from RGB inputs and language instructions, and $\policyNet$ (the \textbf{Diffusion Action Planner}) predicts actions conditioned on $\spatialSignals_t$, raw RGB, and proprioception. This decomposition is central to our design: because SSI is robot-agnostic and requires no action labels, perception modules can be pretrained from diverse video sources, while the downstream policy learns from only a small set of demonstrations.

The \textbf{SSI} encodes spatial structure and task intent using three complementary signals:
(i) monocular depth features $\depthImage$ capturing relative scene geometry,  
(ii) layout maps $\bboxMap$ highlighting task-relevant regions, and  
(iii) instruction-conditioned pixel trajectories $\trajectories = \{ \singleTraj_i \}_{i=1}^N$ indicating anticipated object and manipulator motion.  
Together, these signals convey where relevant objects are, how they relate spatially, and how they are expected to move. Each component is independently computable and trainable from action-free data, supporting modular pretraining and reuse across tasks, scenes, and embodiments.

\subsection{Perception Composer}
The Perception Composer transforms RGB frames and language instructions into the SSI via three independently trained sub-modules: a monocular depth estimator, a language-guided object localizer, and an instruction-conditioned motion predictor. Each operates per-frame without temporal or multi-view fusion, enabling pretraining from large, robot-agnostic datasets.

\paragraph{Monocular Depth Estimation Module (MDE)}
This module infers relative spatial geometry by predicting depth features $\depthImage$ from RGB inputs. Rather than relying on metric depth, which can be noisy and hardware-dependent, we use \textbf{relative depth cues} that capture geometric relationships between the gripper and surrounding objects---sufficient for contact-rich reasoning. We extract intermediate feature maps rather than final depth estimates to improve stability across scenes. We adopt Depth Anything V2~\cite{DepthAnythingV2_Yang_2024} without fine-tuning, while remaining compatible with alternative monocular depth estimators.

\paragraph{Language-Guided Object Localization Module (LGOL)}
To highlight task-relevant regions, this module detects objects and manipulators based on the instruction $\textInstruct$. A language-conditioned object detector proposes bounding boxes, which are aggregated into a dense layout map $\bboxMap$ by filling each region with its confidence score (overlapping areas retain the maximum), yielding a spatial salience map that guides policy attention. An example layout map is shown as an inset in Figure~\ref{fig:fusion_netwrok}.  We use Grounding DINO~\cite{GroundingDINO_Liu_2023} off the shelf; the formulation is detector-agnostic and compatible with finetuned alternatives.

\paragraph{Instruction-Conditioned Motion Prediction Module (ICMP)}
This module complements static cues by forecasting task-relevant motion. Given an RGB frame, instruction, and layout map $\bboxMap$, it predicts a set of pixel trajectories $\singleTraj_i = \{\pixel_i^h\}_{h=1}^{\trajStepLen}$ to model anticipated object and manipulator dynamics.

We introduce a \textbf{hybrid sampling strategy} to initialize trajectories: half of the starting points are sampled uniformly across the image, while the rest are drawn from high-confidence regions in $\bboxMap$, covering both object and non-object regions relevant to the task---a concrete benefit of cross-modal integration within SSI, where the layout map directly informs motion prediction. In contrast, ATM~\cite{ATM_Wen_RSS_2024} uses uniform sampling without task-aware concentration; Im2Flow2Act~\cite{Im2Flow2Act_Xu_CoRL_2024} restricts flow to object-surface keypoints initialized from depth data; and SKIL~\cite{SKIL_wang2025} relies on RGB-D inputs to derive semantic regions. Our approach operates purely in 2D from RGB, requiring no depth sensing or 3D initialization. The predictor architecture builds on ATM's Transformer design, adapted for dense instruction-conditioned trajectory forecasting.

\input{images/fusion_network/fusion_network}

\subsection{DAP: Diffusion Action Planner}
The \textbf{Diffusion Action Planner (DAP)} generates multi-step action sequences conditioned on the Structured Scene Interface (SSI), multi-view RGB inputs, and proprioceptive states. It comprises two stages: multi-modal feature fusion and conditional diffusion-based action generation.

\paragraph{Multi-Modal Feature Fusion}
To construct a unified scene representation for action generation, we integrate spatial and temporal features across modalities (Figure~\ref{fig:fusion_netwrok}). At each observation timestep, RGB images, monocular depth features, and layout maps are encoded per view using modality-specific CNNs. This yields a tensor of shape $\mathbb{R}^{\obsStep \times V \times n_h \times n_w \times 128}$, where $n_h \times n_w$ is the spatial resolution. Features from all modalities are concatenated per view along the channel dimension and projected via a shared MLP into spatial embeddings. These are flattened into tokens and concatenated with trajectory tokens from the motion prediction module. 

A learnable [CLS] token is prepended, and all tokens are enriched with positional and modality-type encodings. The sequence is then processed by a Transformer to perform joint multi-view, multi-modal fusion. The resulting [CLS] tokens across timesteps are concatenated with proprioceptive features, forming the final condition vector $\fusedSpatialFeat_t \in \mathbb{R}^{1 \times (\obsStep \cdot (128 + d_p))}$, where $d_p$ is the proprioceptive feature dimension. We refer to $\fusedSpatialFeat_t$ as the \textit{structured scene vector}, which summarizes the scene context at time $t$ and conditions the diffusion-based action policy.

\paragraph{Action Diffusion}
We model action generation as a conditional denoising diffusion process that transforms Gaussian noise into action sequences, conditioned on the fused scene representation. Given condition vector $\fusedSpatialFeat_t$, the model generates actions $\{a_i\}_{i=t}^{t+H-1}$ by iteratively denoising a noisy initialization.

Following the Diffusion Policy framework~\cite{DP_Chi_RSS_2023}, we train the model to predict noise added during forward diffusion. Let $\mathbf{A}_t^{\denoisingStep}$ denote the noised version of ground-truth action $\mathbf{A}_t$ at denoising timestep $\denoisingStep$. The objective is:
\vspace{-2mm}
\begin{equation}
    \mathcal{L} = \text{MSE}(\epsilon^\denoisingStep - \epsilon_\theta(\fusedSpatialFeat_t, \mathbf{A}_{t}^{\denoisingStep}, \denoisingStep)),
    \vspace{-2mm}
\end{equation}
where $\epsilon^\denoisingStep$ is sampled noise, and $\epsilon_\theta$ is the model’s prediction. At test time, the model samples from a standard Gaussian and performs iterative denoising to produce a coherent action sequence. This generative approach enables the policy to model diverse, multimodal behaviors and improves generalization under limited supervision.
\vspace{-1mm}

%% file: images/overall_framework/item.tex
\begin{figure}[t]
    \centering
    \includegraphics[width=1\linewidth, trim = 80 110 50 210, clip]{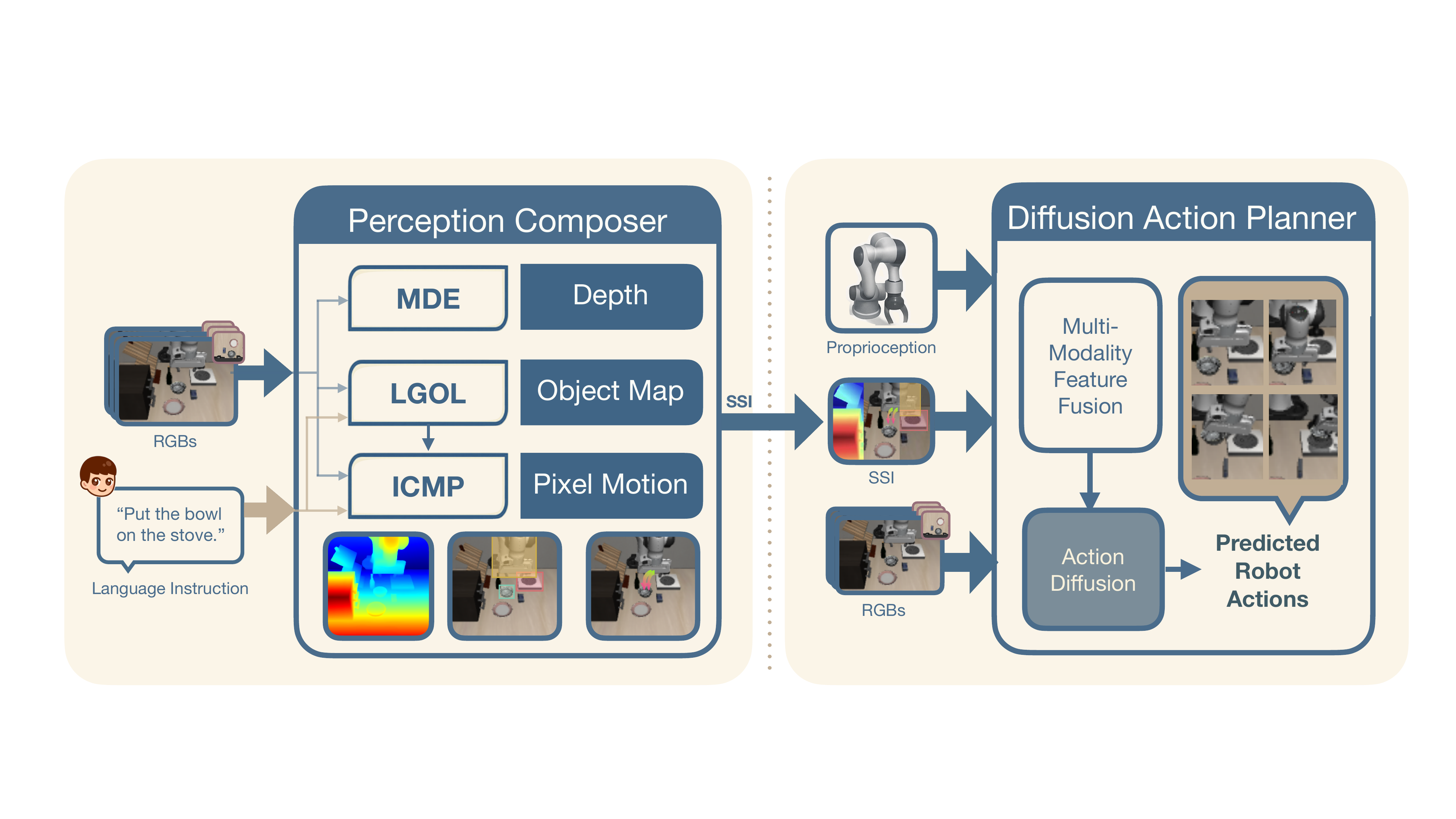}
    \vspace{-26px}
    \caption{\small \textbf{Framework overview.} The Perception Composer converts RGB images and language instructions into three structured signals: monocular depth features, task-relevant layout maps, and instruction-conditioned motion trajectories. The Diffusion Action Planner then integrates these signals—together with proprioception and optional RGB inputs—to generate multi-step action sequences.}
    \label{fig:framework}
\vspace{-16px}

\end{figure}

%% file: images/fusion_network/fusion_network.tex
\begin{figure}[t]
    \centering
    \includegraphics[width=0.8\linewidth]{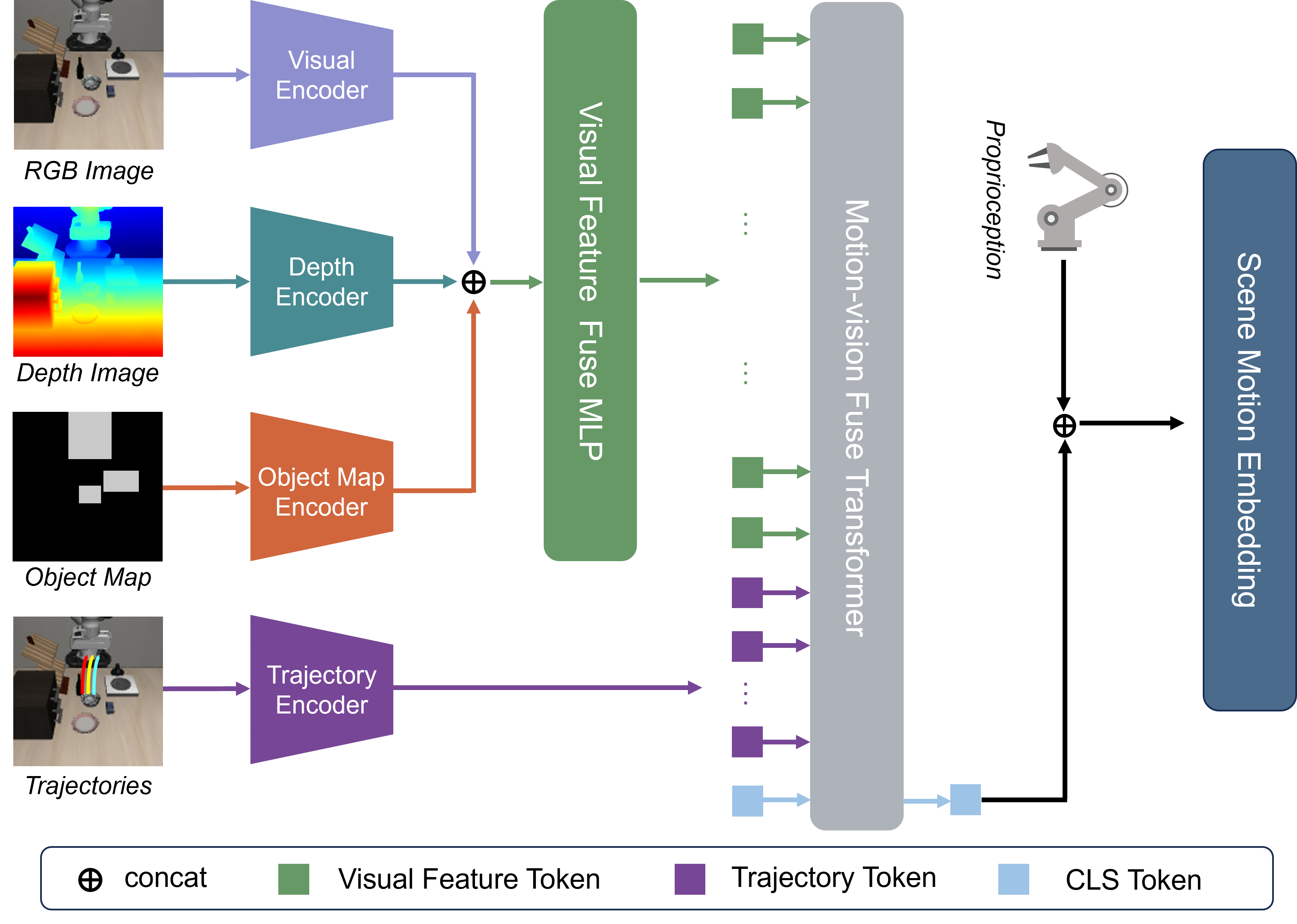}
    \vspace{-5px}
    \caption{\small \textbf{Multi-modal feature fusion architecture.} Per-view visual features (RGB, depth, layout map) are encoded with modality-specific CNNs and fused via an MLP. The resulting tokens, together with motion trajectory tokens, are processed by a Transformer. The output is concatenated with proprioception to form the condition vector for action diffusion.}
    \label{fig:fusion_netwrok}
\vspace{-14px}
\end{figure}

%% file: tex/experiments.tex
\section{Experiment Results}
\label{sec:result}

\vspace{-1mm}
\subsection{Implementation Details}
\vspace{-1mm}
\paragraph{Network Configuration and Training 
Details}
Training and simulation evaluations are performed on four NVIDIA A800 GPUs. Input resolution is $128 \times 128$. We encode language instructions using BERT~\cite{BERT_Devlin_2019} and extract monocular depth features with DepthAnythingV2-ViT-L~\cite{DepthAnythingV2_Yang_2024}. Object bounding boxes are detected by Grounding DINO-T~\cite{GroundingDINO_Liu_2023} with a confidence threshold of 0.35. In simulation, we slightly adjust object prompts to mitigate the visual domain gap caused by low-resolution textures (e.g., replacing ``tomato soup can'' with ``green-orange soup can''). The motion predictor is trained following ATM~\cite{ATM_Wen_RSS_2024} and forecasts 16 future steps for 64 starting pixel points. A 7-layer Transformer (8 heads, hidden dim 128) performs multi-modal fusion. The Diffusion Action Planner follows the U-Net structure of Diffusion Policy~\cite{DP_Chi_RSS_2023}, embedding RGB, depth, and object maps into $8 \times 8$ spatial grids with 128-dimensional tokens. Training uses 100 diffusion steps and AdamW ($\text{lr}=5\times10^{-4}$). The policy is trained for 400 epochs (600 for LIBERO-Long), with image augmentations following ATM~\cite{ATM_Wen_RSS_2024} (color jitter, random shift) and bounding box augmentations (random translation, scaling, deletion, insertion). Unless otherwise specified, we train policies with 10 demonstrations per task, while pretraining the instruction-conditioned motion predictor on all 50 demonstration videos using only RGB observations and language.

\input{images/realworld_platform/item}
\paragraph{Real-world Setup}
We use a 6-DoF ARX-5 arm with two RGB cameras (side-view and eye-in-hand), deployed via ROS at 10~Hz (Figure~\ref{fig:platform_and_tasks}) on a single NVIDIA GeForce RTX 4090. The action space consists of absolute joint and gripper positions, and diffusion inference is performed using 16 denoising steps. We evaluate across 13 tasks covering diverse manipulation capabilities, including spatial reasoning (e.g. directional and relational reasoning), deformable object handling (e.g., sponge, tissue), contact-rich operations (e.g., wiping, drawer closing), multi-stage manipulation, pick-and-place, and cross-embodiment transfer (e.g., human-hand video pretraining). Following ATM~\cite{ATM_Wen_RSS_2024}, we collect 50 teleoperated robot demonstrations per task and 100 human-hand videos for cross-embodiment experiments. Each task is evaluated over 20 rollouts; detailed task descriptions are provided in the corresponding subsections.

\paragraph{Simulation Setup} 
We evaluate on LIBERO~\cite{LIBERO_Liu_2023}, a widely used multi-task manipulation benchmark with standardized task definitions and evaluation protocols. LIBERO features a Panda arm performing language-conditioned tabletop manipulation and contains four suites: \emph{Spatial}, \emph{Goal}, and \emph{Object} (10 tasks each), as well as \emph{LIBERO-100}. The suites emphasize complementary challenges, including spatial relation reasoning (Spatial), goal-conditioned functional behaviors (Goal), and object-centric manipulation and generalization (Object). LIBERO-100 comprises \emph{LIBERO-90} (90 tasks for pretraining) and \emph{LIBERO-10}, a long-horizon suite with 10 multi-stage tasks for evaluation. Each task provides 50 demonstrations with natural-language instructions, multi-view RGB observations, and low-level action labels. We follow ATM~\cite{ATM_Wen_RSS_2024} for the few-shot (10-demonstration) evaluation protocol: episodes run for 600 steps; for each suite, we train a single multi-task policy with three random seeds and evaluate each seed with 20 rollouts per task, reporting the mean success rate across seeds. For diffusion inference, we follow Diffusion Policy~\cite{DP_Chi_RSS_2023} and use 100 denoising steps.

\paragraph{Baselines} 
We compare against a wide range of recent methods. Results for Diffusion Policy~\cite{DP_Chi_RSS_2023} (50 demos) and Octo~\cite{Octo_2024} are taken from OpenVLA~\cite{OpenVLA_Kim_2024}; results for Diffusion Policy (10 demos), R3M~\cite{nair2023r3m}, VPT~\cite{baker2022video}, and UniPi~\cite{UniPi_Du_NIPS_2023} are from ATM~\cite{ATM_Wen_RSS_2024}; remaining baselines use numbers reported in their original papers. When multiple training variants are available, we report the version closest to our setup (e.g., OFT trained only on LIBERO, UniVLA trained on LIBERO with human demonstration enhancement). Some baselines leverage additional large-scale datasets beyond LIBERO for performance gains, whereas our method is trained solely on LIBERO. Our primary quantitative comparisons focus on methods evaluated on LIBERO under RGB-only sensing (no depth sensors; monocular depth is allowed). Approaches that require RGB-D sensing or explicit 3D initialization are discussed in Related Work rather than included as direct baselines.

\input{tables/baseline_10demo}

\vspace{-2mm}
\subsection{Performance Comparison with Existing Methods}
\vspace{-1mm}
We compare against a broad set of baselines including VLA models~\cite{OpenVLA_Kim_2024, VLA_Cache_2025, DiT_Policy_2024, TraceVLA_2024, GRAPE_Zhang_2025, OTTER_2025}, flow-based methods~\cite{FLIP_gao2025, ATM_Wen_RSS_2024}, diffusion policies~\cite{DP_Chi_RSS_2023, Octo_2024}, and spatially grounded models~\cite{DepthHelps_2024_IROS, SpatialVLA_Qu_2025}. Our primary comparisons focus on methods trained solely on LIBERO under comparable sensing assumptions; models leveraging additional large-scale pretraining data (e.g., BridgeV2~\cite{Bridgedata_Walke_CoRL_2023}, OXE~\cite{open_x_embodiment_rt_x_2023}) are included for reference.

\vspace{-1mm}
\paragraph{Low-data regime (main result)}
As shown in Table~\ref{tab:low_data_comparison}, SSI-Policy achieves the best overall performance in the low-data regime, with improvements across all four suites. Compared to the strongest prior method, ATM-DP, our method shows a substantial gain of nearly \textbf{+15\%}. The improvements are especially pronounced on Object (+14.0\%) and Goal (+23.8\%), where structured spatial reasoning and object localization are critical. Notably, the 10-demo SSI-Policy remains competitive with several 50-demo methods that leverage external data (Table~\ref{tab:full_data_comparison}). These results highlight the data efficiency enabled by the SSI: by providing a compact, task-aware interface that captures essential spatial and motion information, the policy can learn effectively from very few demonstrations, reducing the reliance on large-scale pretraining or additional data.

\paragraph{Scalability with more data}
\input{tables/baseline_50demo}
Although SSI-Policy is designed for low-data learning, it scales effectively with additional demonstrations. As shown in Table~\ref{tab:full_data_comparison}, with 50 demonstrations per task, SSI-Policy achieves 91.25\% average success, ranking second among methods trained without external data---just 0.65\% behind OFT~\cite{OFT_2025}. It also remains competitive with models that leverage large-scale pretraining (e.g., SpatialVLA, GRAPE, DiT Policy). Notably, SSI-Policy attains the third-highest Object (97.50\%) and Goal (93.00\%) scores overall, trailing only $\pi_0$ and CLIP-RT, both of which rely on massive external datasets.

\subsection{Spatial Reasoning Capability}
While LIBERO-Spatial evaluates spatial reasoning in simulation, we further assess real-world performance on six tasks: four requiring directional references (e.g., “upper left”) and two requiring disambiguation between identical objects based on spatial relations (e.g., tissue box on the plate vs. on the table). The latter also involves deformable object manipulation, increasing task difficulty. 

We compare against a multi-task Diffusion Policy baseline, a widely used and reproducible RGB-only method for real-world control. For controlled comparison, we reuse the same encoders, architecture, fusion module, and training setup. As shown in Table~\ref{tab:spatial_reasoning}, our method outperforms Diffusion Policy by up to 75 percentage points on directional tasks and 10--20\% on disambiguation tasks (80.0\% vs. 43.3\% on average).

We attribute this to structured spatial grounding: depth cues help resolve directional references, while the SSI combines object localization with motion cues to select the correct instance in cluttered scenes. In contrast, Diffusion Policy frequently confuses relative spatial ordering or selects the incorrect object.

\input{tables/spatial_reasoning}

\subsection{Embodiment Generalization}
\vspace{-0.5mm}

We assess SSI's generalization across embodiments via (1) robot-to-robot transfer in simulation and (2) human-to-robot transfer in real-world settings.

\input{images/experiments/cross_embodiment_full}
\paragraph{Cross-robot transfer} We construct a dataset of 50 videos per task from five non-target robot arms (10 per arm) to train the SSI motion predictor without action labels. The policy is then trained with 10 demonstrations from the target robot, with SSI frozen. We compare \textit{zero-shot} (non-target data only) and \textit{few-shot} (adding 10 target-robot videos) settings, applying the same protocol to ATM for fair comparison. As shown in Figure~\ref{fig:cross_embodiment_full}, our method consistently outperforms ATM in both settings with smaller zero-shot degradation. The few-shot SSI achieves 83.0\%---comparable to the full-shot version (83.5\%)---demonstrating strong embodiment transfer enabled by the SSI abstraction.

\paragraph{Human-hand to robot transfer} 
We further evaluate whether human-hand demonstrations can directly support robot learning, a capability not commonly achieved by existing methods. We test this on two real-world tasks: pick-and-place (e.g., pick up a pot and place it onto the stove) and contact-rich sweeping (e.g., pick up the broom and sweep the toy onto the dustpan). For each task, we collect 100 human-hand videos and 10 robot demonstrations, pretraining SSI's motion predictor on human data and optionally fine-tuning with robot data. The policy is trained using the same 10 robot demos, with SSI kept frozen throughout. As shown in Table~\ref{tab:huam_to_robot}, our method achieves 40--45\% success using human data alone, improving to 70--80\% when combined with robot data---exceeding the robot-only baseline by over 10\%. These results support the claim that SSI provides an embodiment-agnostic abstraction: by separating perception from embodiment-specific control, it enables motion priors learned from human-hand videos to transfer effectively to robot policies.

\input{tables/human2robot}
\input{tables/real_world_comparison}

\subsection{Robustness in Challenging Scenarios}
\vspace{-0.5mm}
To evaluate robustness in complex real-world settings, we design five tasks that are \emph{more challenging} than the earlier settings. Collectively, these tasks span deformable object handling (e.g., sponge), contact-rich interactions (e.g., wiping, drawer closing), long-horizon behaviors, and cluttered scenes with novel distractors. For pick-and-place tasks, we further introduce large variations in object position and orientation to stress spatial generalization.

SSI-Policy is trained as a single multi-task policy across all five tasks and must therefore resolve potential cross-task ambiguity. In contrast, Diffusion Policy~\cite{DP_Chi_RSS_2023} is trained separately for each task, avoiding such ambiguity and thus enjoying a structural advantage. Nevertheless, as shown in Table~\ref{tab:real_comparison}, SSI-Policy consistently outperforms the baseline (54\% vs. 29\% average success), demonstrating robust performance under challenging conditions.

\subsection{Ablation Studies}
\vspace{-0.5mm}
We conduct ablation studies on LIBERO to quantify the contribution of each component in SSI-Policy and to assess the importance of jointly modeling geometry and task-conditioned motion. All variants are trained with 10 demonstrations per task, and results are averaged over three random seeds.

\input{tables/ablation}
\input{images/experiments/SSI_results}

\paragraph{Component ablations and complementarity}
Table~\ref{tab:ablation} shows that the full model achieves the best overall performance (80.42\%). Geometry-only (A0) and motion-only baselines (A1--A2) perform substantially worse, especially on Long-horizon tasks, indicating that neither cue alone is sufficient. Adding depth to motion (A3) yields the strongest Spatial score (85.83\%), while adding layout maps (A4) improves Goal tasks (81.50\%); however, these partial variants are specialized and do not generalize across suites. The unified interface performs consistently well across suites, indicating complementary benefits from jointly encoding geometric structure and task-conditioned motion within a shared representation. Removing hybrid sampling (VAR) reduces the average from 80.42\% to 78.46\%, showing a consistent but incremental benefit. 

\paragraph{Influence of SSI}
To further isolate the role of SSI, we train a variant conditioned only on SSI and proprioception, excluding raw RGB input for the diffusion action planner. As shown in Figure~\ref{fig:SSI_results}, this SSI-only policy achieves 79.3\%, retaining $\sim$98\% of the full model's 80.4\%. This confirms that the structured interface captures the majority of task-relevant spatial and motion information rather than serving as auxiliary features.

\vspace{-0.8mm}
\subsection{Qualitative Results}
\vspace{-1mm}
We present visualizations of successful executions in both LIBERO and real-world settings in Figure~\ref{fig:qualitative_LIBERO}.

\vspace{-0.5mm}
\subsection{Failure Cases}
\vspace{-1mm}
While \textbf{SSI-Policy} performs well overall, we observe three recurring failure modes. First, object detection errors under clutter or occlusion can lead to missed or inaccurate GroundingDINO boxes, corrupting the layout map and downstream decisions. Second, in tasks involving deformable or bulky objects (e.g., sponges), grasp failures can occur due to object displacement during execution, exacerbated by limited gripper aperture and friction. Third, temporal inconsistency in monocular depth or detections can induce jittery motion predictions. Overall, most failures originate from upstream perception noise, suggesting the structured interface is reliable when perceptual inputs are stable. These observations motivate future work on stronger detection backbones, temporally consistent perception, and improved physical interaction via compliance or tactile feedback.

\input{images/qualitative_LIBERO/item}

%% file: images/realworld_platform/item.tex
\begin{figure}[!t]
    \centering
    \includegraphics[width=1\linewidth, trim = 152 180 730 125, clip]{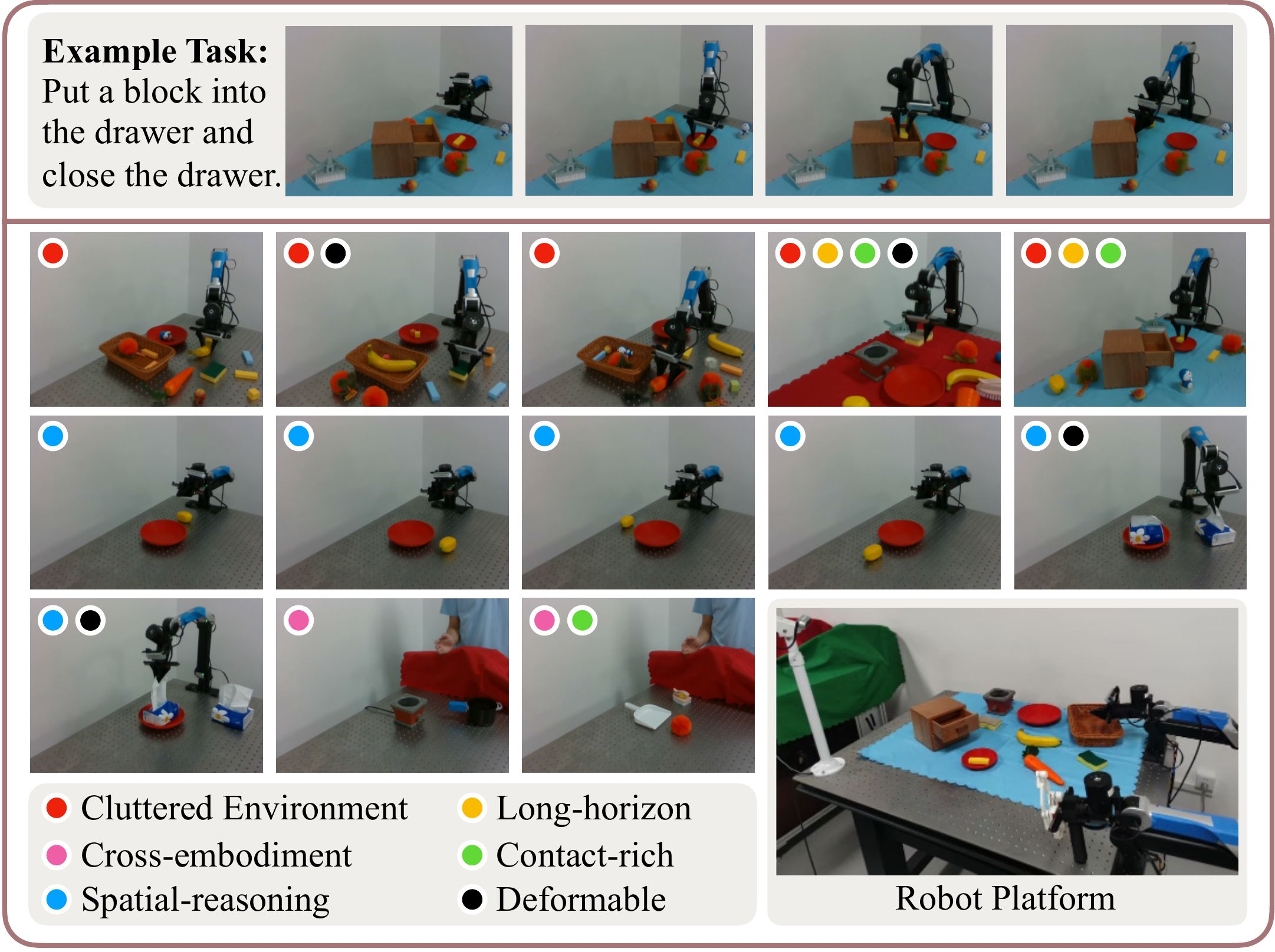}
    \vspace{-17px}
    \caption{\small \textbf{Real-world setup and task suite.} 
    \textbf{Top:} time-lapse of a representative task. 
    \textbf{Middle/Bottom-left:} representative scenes illustrating the diversity of the 13 real-world tasks, annotated with colored markers indicating task categories. 
    \textbf{Bottom-right:} full view of the experimental platform, including the 6-DoF robotic arm and dual RGB cameras (side-view and eye-in-hand).}
    \label{fig:platform_and_tasks}
\vspace{-15px}
\end{figure}

%% file: tables/baseline_10demo.tex
\begin{table}[t]
    \caption{\small \textbf{Few-shot (10 demonstrations) performance on LIBERO.} Success rates (\%) across four task suites. 
    }
    \vspace{-1.5mm}
    \tabcolsep = 3.0pt
    \centering
    \renewcommand{\arraystretch}{1.0}
    \resizebox{1\linewidth}{!}
    {
    \footnotesize
    \begin{tabular}{@{}l|c|c|c|c|c@{}}
        \toprule[0.4mm]
        \textbf{Method} & \textbf{Spatial}$\uparrow$ & \textbf{Object}$\uparrow$ & \textbf{Goal}$\uparrow$ & \textbf{Long}$\uparrow$ & \textbf{Average}$\uparrow$\\
        \midrule[0.2mm]
        \midrule[0.2mm]
        R3M-finetune~\cite{nair2023r3m} & 49.17 & 52.83 & 5.33 & 9.17 & 29.13\\
        VPT~\cite{baker2022video} & 37.83 & 19.50 & 3.33 & 3.83 & 16.13\\
        UniPi~\cite{UniPi_Du_NIPS_2023} & 69.17 & 59.83 & 11.83 & 5.83 & 36.67\\
        Diffusion Policy~\cite{DP_Chi_RSS_2023} & 67.67 & 78.00 & 35.00 & 37.33 & 54.50\\
        ATM~\cite{ATM_Wen_RSS_2024} & 68.50 & 68.00 & \underline{77.83} & 39.33 & 63.42\\
        ATM-DP~\cite{ATM_Wen_RSS_2024} & \underline{79.00} & {81.00} & 58.67 & 44.00 & \underline{65.67}\\
        MaIL~\cite{MaIL_Jia_2024} & 60.90 & \underline{85.40} & 56.30 & {46.00} & 62.15\\
        FLIP\tablefootnote{Approximate value estimated from Figure 5 of FLIP (ICLR 2025). The paper reports results only on LIBERO-LONG (10 tasks) and does not provide exact numeric values or per-suite breakdowns.}~\cite{FLIP_gao2025} & - & - & - & \underline{$\approx$50} & -\\
        \midrule
        \textbf{SSI-Policy (Ours)} & \textbf{83.50} & \textbf{95.00} & \textbf{82.50} & \textbf{60.67} & \textbf{80.42}\\
        \bottomrule
    \end{tabular}
    }
    \label{tab:low_data_comparison}
    \vspace{-14px}
\end{table}

%% file: tables/baseline_50demo.tex
\begin{table}[t]
    \caption{\small \textbf{Scalability with more demonstrations on LIBERO (50 demos per task).}
    Success rates (\%) across four suites. 
    Bold/underline: 1st/2nd (no external data).
    ``Ext. Data'' denotes additional large-scale datasets (e.g., BridgeV2, OXE). 
    External-data methods and single-view variants (*) are included for completeness.}
    \vspace{-1.5mm}
    \tabcolsep = 2.5pt
    \centering
    \renewcommand{\arraystretch}{1.05}
    \resizebox{1.0\linewidth}{!}
    {
    \footnotesize
    \begin{tabular}{@{}l|c|c|c|c|c|c@{}}
        \toprule[0.4mm]
        \textbf{Method} & \textbf{Ext. Data} & \textbf{Spatial}$\uparrow$ & \textbf{Object}$\uparrow$ & \textbf{Goal}$\uparrow$ & \textbf{Long}$\uparrow$ & \textbf{Average}$\uparrow$\\
        \midrule[0.2mm]
        \midrule[0.2mm]
        Diffusion Policy*~\cite{DP_Chi_RSS_2023} & & 78.30 & 92.50 & 68.30 & 50.50 & 72.40\\
        MDT~\cite{MDT_Reuss_RSS_2024} & & 78.50 & 87.50 & 73.50 & 64.80 & 76.10\\
        DepthHelps~\cite{DepthHelps_2024_IROS} & & 71.20 & 78.60 & 66.40 & 36.40 & 63.15\\
        OFT~\cite{OFT_2025} & & \textbf{94.30} & \underline{95.20} & \underline{91.70} & \textbf{86.50} & \textbf{91.90}\\
        \midrule
        \textbf{SSI-Policy (Ours)} & & \underline{94.00} & \textbf{97.50} & \textbf{93.00} & \underline{80.50} & \underline{91.25}\\
        \midrule[0.3mm]
        \rowcolor{gray!20}
        \multicolumn{7}{@{}l}{{\textit{Methods with additional pretraining data (for reference)}}}\\
        \midrule[0.1mm]
        Octo*~\cite{Octo_2024} & \cmark & 78.90 & 85.70 & 84.60 & 51.10 & 75.20\\
        OpenVLA*~\cite{OpenVLA_Kim_2024} & \cmark & 84.70 & 88.40 & 79.20 & 53.70 & 76.50\\
        SpatialVLA~\cite{SpatialVLA_Qu_2025} & \cmark & 88.20 & 89.90 & 78.60 & 55.50 & 78.10\\
        VLA-Cache~\cite{VLA_Cache_2025} & \cmark & 83.80 & 85.80 & 76.40 & 52.80 & 74.70\\
        TraceVLA~\cite{TraceVLA_2024} & \cmark & 84.60 & 85.20 & 75.10 & 54.40 & 74.80\\
        OTTER~\cite{OTTER_2025} & \cmark & 84.00 & 89.00 & 82.00 & - & -\\
        GRAPE~\cite{GRAPE_Zhang_2025} & \cmark & 88.50 & 92.10 & 83.10 & 57.20 & 80.20\\
        DiT Policy~\cite{DiT_Policy_2024} & \cmark & 84.20 & 96.30 & 85.40 & 63.80 & 82.40\\
        FAST~\cite{FAST_RSS_25} & \cmark & 96.40 & 96.80 & 88.60 & 60.20 & 85.50\\
        UniVLA\tablefootnote{Full LIBERO + human demonstration pretraining.}~\cite{UniVLA_RSS_25} & \cmark & 91.20 & 94.20 & 90.20 & 79.40 & 88.70\\
        CLIP-RT~\cite{CLIP-RT_RSS_25} & \cmark & 95.20 & 99.20 & 94.20 & 82.60 & 92.80\\
        $\pi_0$~\cite{pi0_Black_2024} & \cmark & 96.80 & 98.80 & 95.80 & 85.20 & 94.20\\
        \bottomrule
    \end{tabular}
    }
    \label{tab:full_data_comparison}
    \vspace{-13px}
\end{table}

%% file: tables/spatial_reasoning.tex
\begin{table}[t]
\centering
\caption{\small \textbf{Real-world spatial reasoning evaluation.} Success rates (\%) comparing our method to Diffusion Policy (DP) on six tasks. Tasks include directional reasoning and disambiguation between identical objects based on spatial context.}
\vspace{-1.5mm}
\footnotesize
\resizebox{1\linewidth}{!}{
\begin{tabular}{@{}l|c|c@{}}
\toprule
\textbf{Task} & \textbf{DP} & \textbf{Ours} \\
\midrule
\midrule
Place the lemon \textbf{upper right} of the red plate onto the plate & 55.00 & \textbf{85.00}\\
Place the lemon \textbf{upper left} of the red plate onto the plate & 40.00 & \textbf{80.00}\\
Place the lemon \textbf{lower right} of the red plate onto the plate & 55.00 & \textbf{100.00}\\
Place the lemon \textbf{lower left} of the red plate onto the plate & 10.00 & \textbf{85.00}\\
\midrule
Take out a tissue from the \textbf{tissue box on the plate}& 50.00 & \textbf{60.00}\\
Take out a tissue from the \textbf{tissue box on the table}& 50.00 & \textbf{70.00}\\
\midrule
\textbf{Average} & 43.33 & \textbf{80.00}\\
\bottomrule
\end{tabular}}
\vspace{-2.5mm}
\label{tab:spatial_reasoning}
\end{table}

%% file: images/experiments/cross_embodiment_full.tex
\begin{figure}[t]
    \centering
    \includegraphics[width=0.58\linewidth, trim = 10 12 10 35, clip]{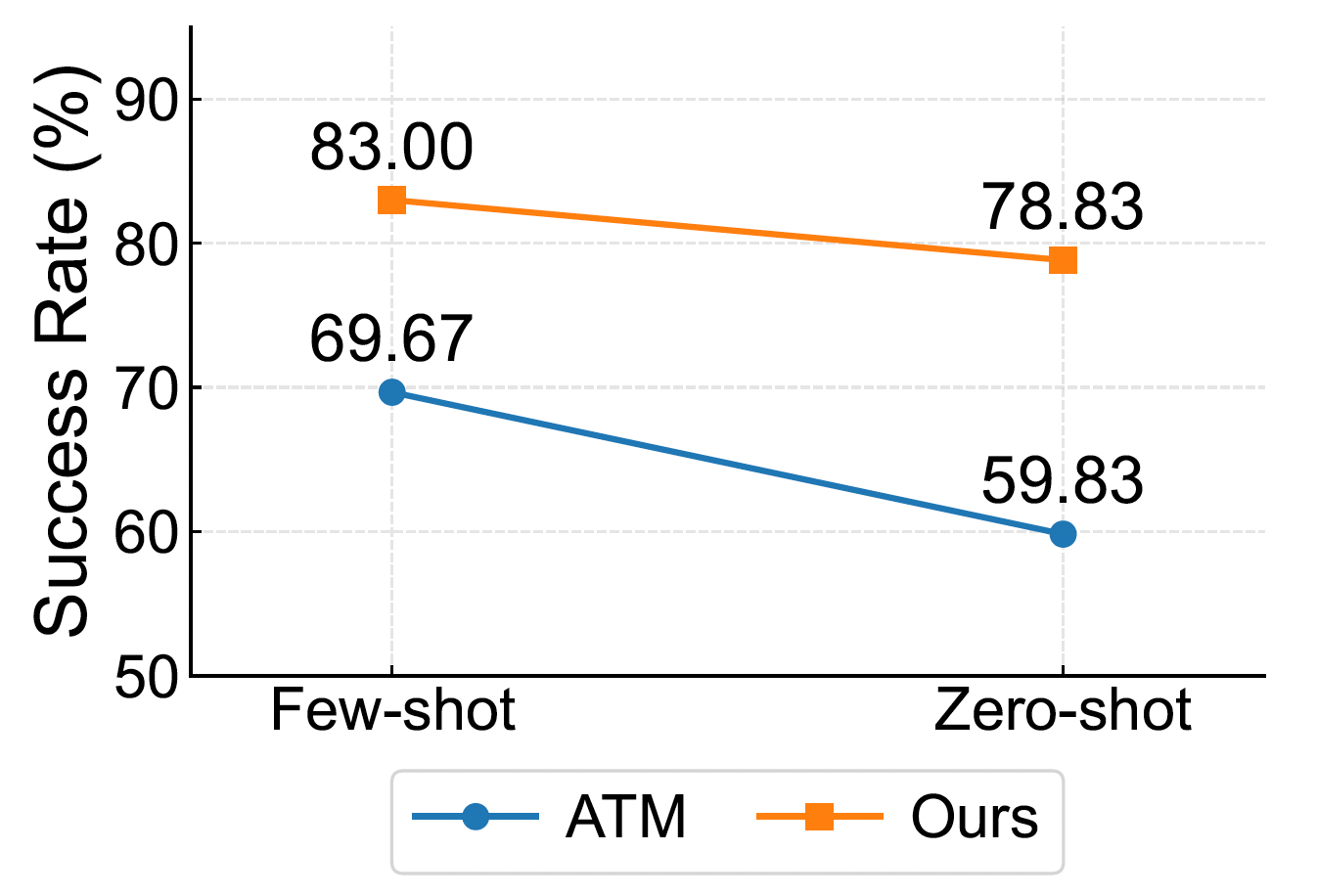}
    \vspace{-1.5mm}
    \caption{\small \textbf{Cross-embodiment results on LIBERO-Spatial.} Success rates under few-shot and zero-shot settings. 
    }
    \label{fig:cross_embodiment_full}
    \vspace{-13px}
\end{figure}

%% file: tables/human2robot.tex
\begin{table}[t]
\centering
\caption{\small \textbf{Human-to-robot transfer.} Success rates (\%) with SSI trained on human-hand data, robot data, or both.}
\vspace{-1mm}
\footnotesize
\resizebox{1\linewidth}{!}{
\begin{tabular}{@{}l|c|c|c@{}}
\toprule
\textbf{Task} & \textbf{Human} & \textbf{Robot} & \textbf{Human+Robot} \\
\midrule
\midrule
Pick up a pot and place it onto the stove & 45.00 &  60.00 & \textbf{70.00}\\
Pick up the broom and sweep the toy onto the dustpan  & 40.00 & 70.00    & \textbf{80.00} \\
\bottomrule
\end{tabular}}
\label{tab:huam_to_robot}
\vspace{-2mm}
\end{table}

%% file: tables/real_world_comparison.tex
\begin{table}
    \centering
    \renewcommand{\arraystretch}{1.0}
    \tabcolsep=5pt
    \caption{\small \textbf{Robustness on challenging real-world tasks.} Success rates (\%) comparing our multi-task policy to single-task Diffusion Policy (DP).}
    \label{tab:real_comparison}
    \vspace{-1.5mm}
    \resizebox{1\linewidth}{!}
    {
    \footnotesize
    \begin{tabular}{@{}l|c|c@{}}
        \toprule[0.4mm]
        \textbf{Task} & \textbf{DP} & \textbf{Ours} \\
        \midrule[0.2mm]
        \midrule
        \text{Put the sponge onto the plate} & 10.00 & \textbf{65.00} \\
        \text{Put the carrot into the basket} & 30.00 & \textbf{75.00} \\
        \text{Put a block into the drawer and close the drawer} & 25.00 & \textbf{40.00} \\
        \text{Put the banana into the basket} & 45.00 & \textbf{50.00} \\
        \text{Pick up a sponge, wipe the table and put the sponge onto the plate} & 35.00 & \textbf{40.00} \\
        \midrule
        \textbf{Average} & 29.00 &  \textbf{54.00}\\
        \bottomrule
    \end{tabular}}
    \vspace{-4mm}
\end{table}

%% file: tables/ablation.tex
\begin{table}[t]
    \caption{\small \textbf{Ablation studies on LIBERO (10 demos per task).} M: motion trajectories, D: monocular depth, L: layout map. 
    }
    \label{tab:ablation}
    \vspace{-1.5mm}
    \tabcolsep = 2.0pt
    \centering
    \renewcommand{\arraystretch}{1.05}
    \resizebox{1.0\linewidth}{!}
    {
    \footnotesize
    \begin{tabular}{@{}l|c|c|c|c|c|c|c|c@{}}
        \toprule[0.4mm]
        \textbf{Variant} & \textbf{M} & \textbf{D} & \textbf{L} & \textbf{Spatial}$\uparrow$ & \textbf{Object}$\uparrow$ & \textbf{Goal}$\uparrow$ & \textbf{Long}$\uparrow$ & \textbf{Avg.}$\uparrow$\\
        \midrule[0.2mm]
        \midrule[0.2mm]
        \textbf{SSI-Policy (Ours)} & \cmark & \cmark & \cmark & {83.50} & \textbf{95.00} & \textbf{82.50} & \textbf{60.67} & \textbf{80.42}\\
        \midrule
        (A0) DepthHelps~\cite{DepthHelps_2024_IROS} & \xmark & \cmark & \xmark & 71.20 & 78.60 & 66.40 & 36.40 & 63.15\\
        (A1) ATM~\cite{ATM_Wen_RSS_2024} & \cmark & \xmark & \xmark & 68.50 & 68.00 & 77.83 & 39.33 & 63.42\\
        (A2) ATM-DP~\cite{ATM_Wen_RSS_2024} & \cmark & \xmark & \xmark & 79.00 & 81.00 & 58.67 & 44.00 & 65.67\\
        (A3) Motion + Depth & \cmark & \cmark & \xmark & \textbf{85.83} & 87.83 & 76.33 & 57.67 & 76.92\\
        (A4) Motion + Layout map & \cmark & \xmark & \cmark & 83.00 & 89.67 & 81.50 & 54.17 & 77.09\\
        \midrule
        (VAR) w/o hybrid sampling & \cmark & \cmark & \cmark & 80.83 & 94.17 & 79.33 & 59.50 & 78.46\\
        \bottomrule
    \end{tabular}
    }
    \vspace{-10px}
\end{table}

%% file: images/experiments/SSI_results.tex
\begin{figure}[t]
    \centering
    \vspace{-0.2mm}
    \includegraphics[width=0.87\linewidth]{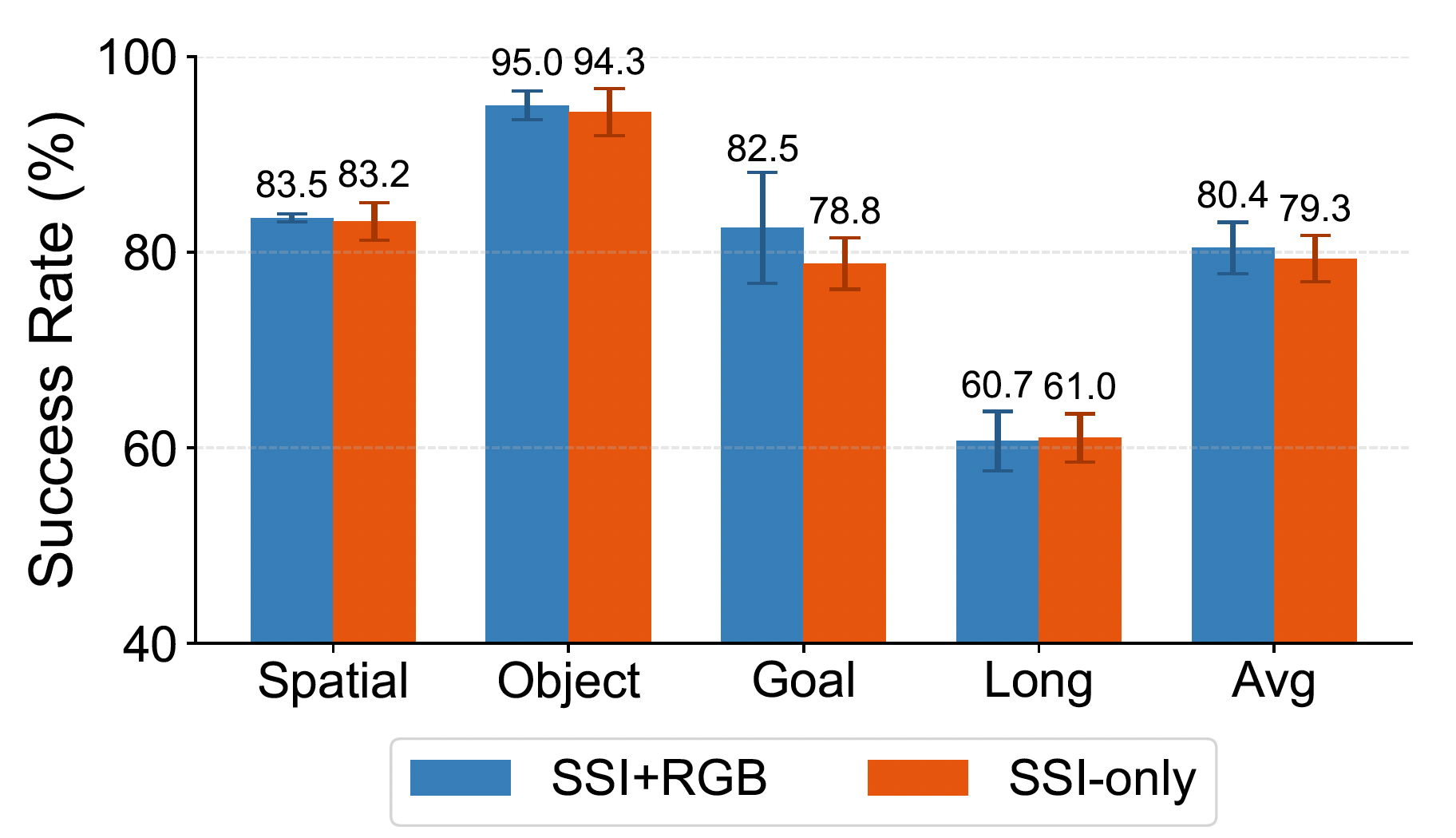}
    \vspace{-7px}
    \caption{\small \textbf{Evaluating SSI as a policy interface.} Success rates across LIBERO suites comparing the full model (SSI + RGB) with an SSI-only variant conditioned on SSI and proprioception.
    } 
    \label{fig:SSI_results}
    \vspace{-17px}
\end{figure}

%% file: images/qualitative_LIBERO/item.tex
\begin{figure}[!t]
    \centering
    \includegraphics[width=1\linewidth, trim = 0 0 0 0, clip]{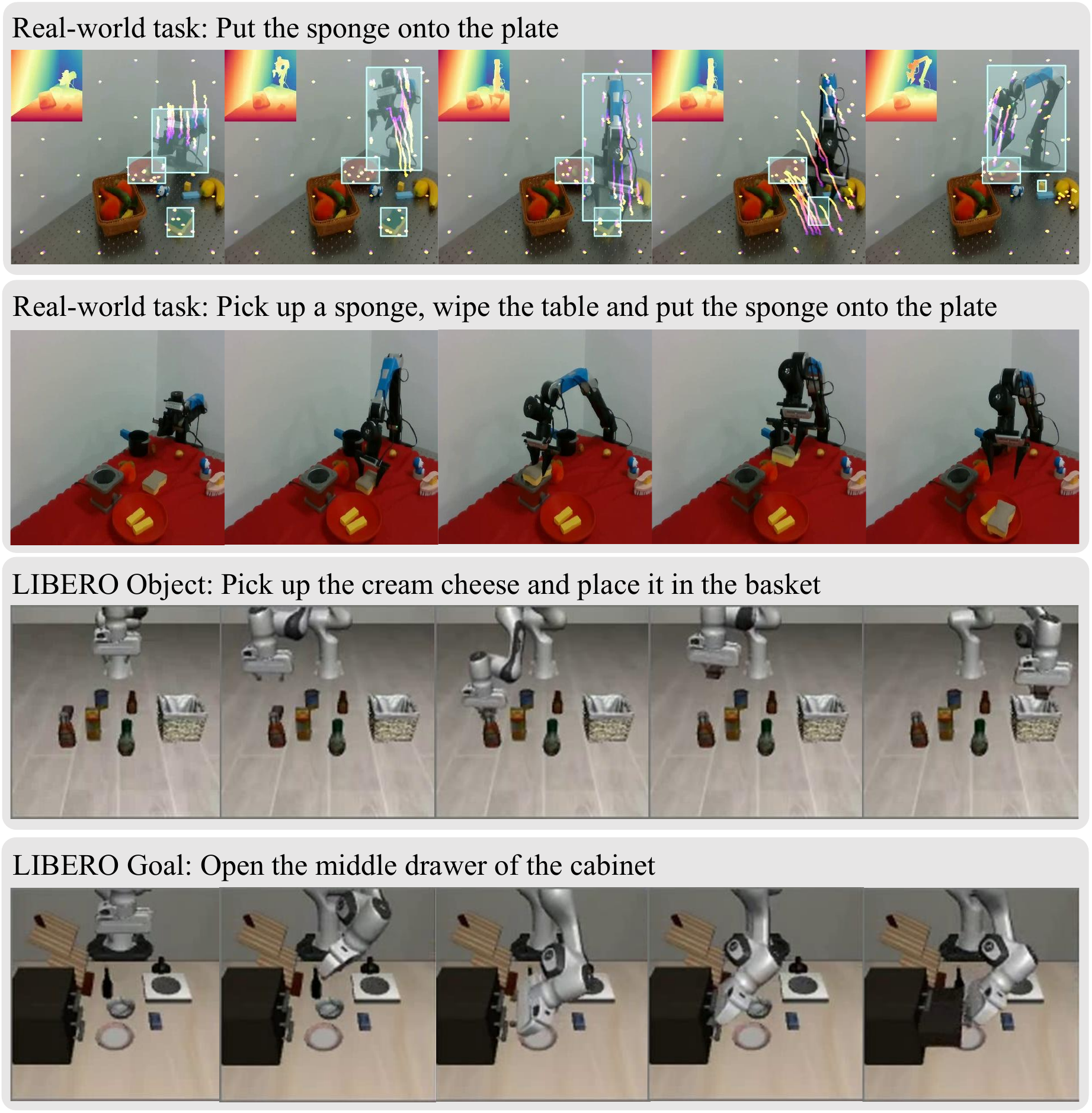}
    \vspace{-6mm}
    \caption{\small 
    \textbf{Qualitative snapshots of task executions.} Each row shows frames from a single rollout. Top two rows: real-world experiments (with SSI overlays in the first row). Bottom two rows: simulated LIBERO tasks. 
}
    \label{fig:qualitative_LIBERO}
\vspace{-15px}
\end{figure}

%% file: tex/conclusion.tex
\vspace{-2mm}
\section{Conclusion}
\label{sec:conclusion}
We presented SSI-Policy, a framework centered on a Structured Scene Interface (SSI)---a unified, RGB-only intermediate representation for vision-language manipulation. By jointly encoding monocular geometry, language-grounded layouts, and instruction-conditioned motion, SSI provides a robot-agnostic abstraction that decouples perception from embodiment-specific control---enabling pretraining from action-free video and policy learning from as few as 10 demonstrations per task. Ablations confirm that geometric and motion cues are complementary within the shared interface. Experiments on LIBERO and 13 real-world tasks demonstrate strong few-shot performance, cross-embodiment transfer, and robustness in contact-rich settings, without depth sensors or external pretraining data. Future work will address temporally consistent perception and tactile feedback for more robust manipulation.

\vspace{-2mm}